\documentclass{article} 
\usepackage{iclr2027_conference,times}

\usepackage[T1]{fontenc}
\usepackage[utf8]{inputenc}
\usepackage{microtype}
\usepackage{inconsolata}

\usepackage{graphicx}
\usepackage{subcaption}
\usepackage{amsmath}
\usepackage{amsfonts}
\usepackage{amssymb}
\usepackage{enumerate}
\usepackage{booktabs}

\usepackage{xcolor}
\usepackage{xparse}
\usepackage{etoolbox}
\usepackage{mdframed}

\usepackage{amsmath,amsfonts,bm}

\def\eqref#1{equation~\ref{#1}}

\def\1{\bm{1}}

\DeclareMathAlphabet{\mathsfit}{\encodingdefault}{\sfdefault}{m}{sl}
\SetMathAlphabet{\mathsfit}{bold}{\encodingdefault}{\sfdefault}{bx}{n}

\usepackage{hyperref}
\usepackage{url}

\usepackage{float}

\newcommand{\set}[1]{\left\{#1\right\}}
\newcommand{\tuple}[1]{\left\langle #1\right\rangle}
\newcommand{\ap}[1]{\textcolor{crayola-bright}{[AP: #1]}}

\definecolor{crayola-bright}{HTML}{EEB14F}
\definecolor{shimmer-bright}{HTML}{B22222}

\NewDocumentCommand{\child}{mmm}{%
   \ensuremath{
      Y%
      \ifstrempty{#1}{}{_{#1}}%
         \ifstrempty{#2}{%
               \ifstrempty{#3}{}{^{#3}}%
               }{%
                  \ifstrempty{#3}{^{#2}}{^{#2,#3}}%
                  }%
      }
}
\NewDocumentCommand{\parent}{mm}{%
   \ensuremath{
      X%
      \ifstrempty{#1}{}{_{#1}}%
      \ifstrempty{#2}{}{^{#2}}%
   }
}

\newmdenv[
   topline=false,
   bottomline=false,
   rightline=false,
   linecolor=gray,
   linewidth=1pt,
   skipabove=\topsep,
   skipbelow=\topsep
]{draftytext}

\title{Small transformers track Bayesian evidence for latent common causes via a context-invariant mechanism}

\author{Amir Mohammadpour \& Michael Franke \\
 Department of Linguistics, University of T\"ubingen \\
 \texttt{a.mohammad-pour@uni-tuebingen.de}
}

\iclrfinalcopy 
\begin{document}

\maketitle
\lhead{Preprint. Under review.}

\begin{abstract}
  We present an in-depth investigation of how a form of Bayesian reasoning about common causes can emerge as a cross-contextual generalization in small, tractable transformers.
  Incrementing on recent work, our set-up (i) disentangles causal mechanisms in the model from the causal structure of the true data-generating process, (ii) orients more towards natural language prediction by considering inference of latent common causes, and (iii) considers  whether and how Bayesian evidence accumulation for latent common causes can be implemented in representations and mechanisms that allow for cross-context generalization to novel test cases.
\end{abstract}

\section{Introduction}

Given the performance of large transformers, it is likely that these models acquire implicit world models which emerges as useful compressions in service of accurate prediction \citep[e.g.,][]{Yudkowsky2023:GPTs-are-Predic,BereskaGavves2024:Mechanistic-Int,shai2026transformerslearnfactoredrepresentations}.
Evidence comes from case studies on finite-state problems, such as board games \citep{ToshniwalWiseman2022:Chess-as-a-Test,LiHopkins2023:Emergent-World-}, taxi driving routes \citep{VafaChen2024:Evaluating-the-}, or text-based games \citep{LiNye2021:Implicit-repres}.
While it remains unclear how pressure for sequence prediction leads to the emergence of world models \citep{li2025doesmeanneuralnetwork,yuan2025revisitingothelloworldmodel}, it is plausible that world models support the ability to perform probabilistic causal reasoning.

\begin{figure}
    \centering
    \includegraphics[width=\linewidth]{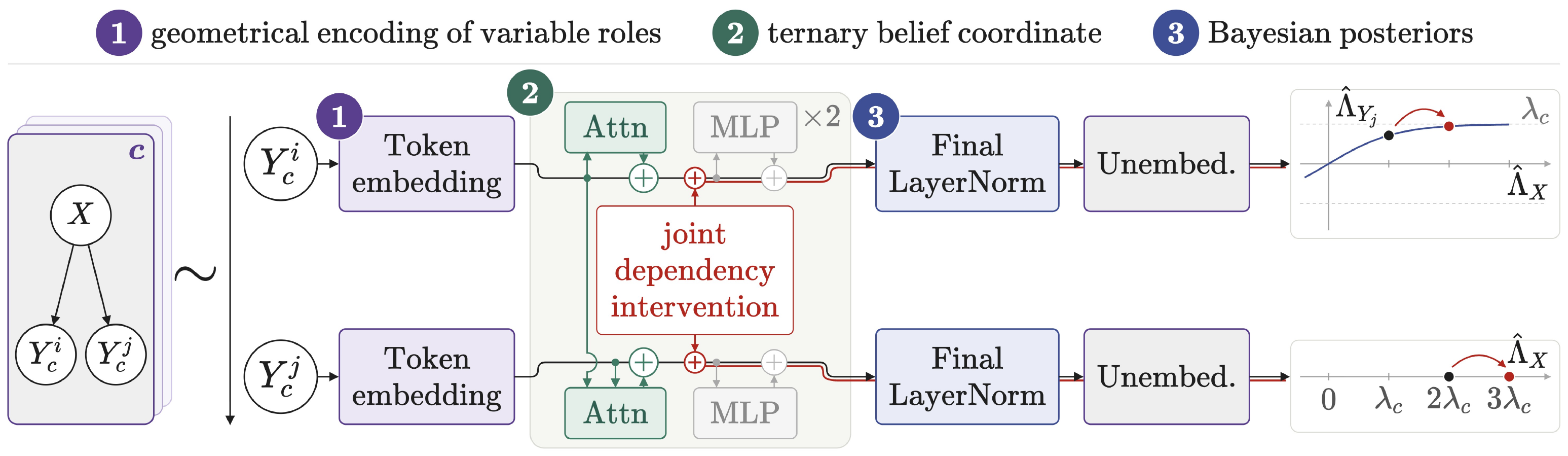}
    \caption{
        We train small transformers on scrambled token sequences from different contexts, each of which are causal models with a common cause and two effect variables. 
        Held-out contexts require the models to extrapolate Bayesian inference of the causal variable's value from contexts seen in training.
        We find that ``Bayesian generalization'' emerges, and we detail its mechanics, finding that 
        (1) token embeddings geometrically encode the relevant roles of the underlying causal variables, 
        that (2) the residual stream encodes a categorical belief representation, 
        which is (3) transformed via the LayerNorm into an approximate Bayesian posterior.
        Surgical interventions yield systematic downstream effects on the joint distribution over relevant variables, as predicted by Bayes rule.
        }
    \label{fig:results-overview} 
\end{figure}

This paper seeks to contribute to the investigation of emergence of world models by an in-depth, mechanistic analysis of generalizable Bayesian (causal) reasoning emerging in a novel, minimalistic experimental setup.
Conceptually, we contribute an analytical delineation of different senses of causal information relevant for the analysis of language models and adopt an architecturally aware \textit{ideal-learner analysis}, which supports investigating how Bayesian abductive reasoning (from observed effects to underlying causes) can emerge in a held-out context by exploiting similarity in the representations of the abstract conceptual roles that different tokens play \citep[e.g.,][]{Shepard1987:Towards-a-Unive,wang2026mathematicaltheoryunderstandingabstract}.
We argue that this kind of cross-context Bayesian generalization amounts to a simple form of \emph{causal representation learning}, i.e., recovery of relevant causal variables, but we also stress the general theoretical limits of causal recovery \citep{spirtes2000causation,richardson2002ancestral}.
Empirically, we demonstrate how small transformers indeed show a form of cross-context ``Bayesian generalization'' in their behavior
by showing that generalizable Bayesian-like behavior is indeed supported by internal ``Bayes-like'' representations, the systematic manipulation of which with suitable \textit{joint-dependency interventions} leads to downstream effects that are coherent with Bayesian computations, thus satisfying the foundational abstraction requirement of commutativity under intervention \citep{RubensteinWeichwald2017:Causal-Consiste,BeckersHalpern2019:Abstracting-Cau,GeigerLu2021:Causal-Abstract,XiaBareinboim2024:Neural-Causal-A}.
Our methods\footnote{Code for all the experiments in \href{https://anonymous.4open.science/r/bayformer-paper-19FF/}{https://anonymous.4open.science/r/bayformer-paper-19FF/}} and main results are succinctly summarized in Figure~\ref{fig:results-overview}. 

\section{Related work \& novel contributions}

\paragraph{Causation.}
For large transformers one important area of research is \emph{represented causality}, i.e., whether models process token sequences which contain information about causality (at least to human observers) in a way that is correct \citep{JinChen2023:CLadder:-A-Benc,KicimanNess2024:Causal-Reasonin,ChiLi2024:Unveiling-Causa} or human-like \citep{BinzSchulz2023:Using-cognitive}.
Smaller transformers have been analyzed in-depth with hypothesis-driven experiments about causality and its representation and relevance.
For example, \citet{RohekarGurwicz2023:Causal-Interpre} show that a transformer’s self-attention allocation can be used for estimating a structural causal model of the transformer’s internal computational mechanism, and \citet{nichani2024transformers} argue based on emerging attentional mechanisms that transformers can learn the true causal processes that generated the data.
Emerging attention patterns track these position-structural dependencies and these, by design, happen to coincide with the true data-generating process.
In both cases, the assumed data-generating process is a simple Markov process in which the token at position $i$ depends only on the token at a prior position $i-k$, so attention patterns tracking these positional dependencies coincide with the true process by design.
As a results, these experimental designs, while doubtlessly insightful to a fair extent, entangle an important distinction between what we here call \textit{within-model causality}, which involves causal mechanisms operating in the model (abstract properties of the forward pass computation), and \emph{behind-data causality}, which concerns information about causal processes that occur outside of the model (in the data-generating process) but which is nevertheless represented and functionally relevant in the model.
Clearly separating \textit{within-model causality} from \emph{behind-data causality} is important, because full natural language generation is a arguably a latent variable problem, in which the next token prediction may depend on abstract, latent variables for which prior tokens at variable positions provide only indirect probabilistic evidence \citep[e.g.,][]{Jiang2023:A-Latent-Space-,ZhangMcCoy2023:Deep-de-Finetti}.
This is why our experimental setup uses a data-generating process in which there are (partially latent, partially observed) common causes for overt tokens, and where surface-form realization is shuffled to be more position-invariant.

\paragraph{Bayes.}

While the connection between probabilistic machine learning and (normatively correct) Bayesian inference is an obvious foundational issue \citep[e.g.,][]{MacKay:2003_InformationTheory}, a recent prominent topic is whether in-context learning in large language models constitutes a form of Bayesian inference \citep[e.g,][]{XieRaghunathan2022:An-Explanation-,raventós2023pretrainingtaskdiversityemergence,falck2024incontextlearninglargelanguage,CaoHe2025:Transformers-Si}.
Other work investigated whether Bayesian / causal reasoning abilities emerge that are normatively correct or human-like \citep[e.g.,][]{ShwartzChoi2020:Do-Neural-Langu,muller2021transformers,KaufIvanova2022:Event-knowledge,ZhuGriffiths2024:Incoherent-Prob,LiRui2024:Dual-Traits-in-,GuptaCorona2025:Enough-Coin-Fli}.
There is also work that explores specific training regimes that induce or enhance Bayesian reasoning abilities on transformers \citep[e.g.,][]{HuJain2024:Amortizing-intr,QiuSha2026:Bayesian-teachi}.

This paper primarily asks how transformers could implement generalizable Bayesian reasoning, contributing to several recent strands of related research.
On controlled synthetic tasks, \citet{AgarwalDalal2026:The-Bayesian-Ge} find attention consistent with incremental Bayesian posterior computation along the residual stream.
Relatedly, a line of work inspired by \textit{computational mechanics}\citep{shaliziComputationalMechanicsPattern2001},
trace the geometry of latent (Bayesian) beliefs in the residual stream \citep[][]{shai2024transformers, levinson2026finding}, and further argue that transformers implement a constrained sequential belief update \citealp[]{piotrowski2025constrained}.
So far, however, these studies have focused on cases where transformers' predictions and internal computations are analyzed on cases that were included in the training sets, and causal-interventionist checks on recovered representations have not always been given.

\paragraph{Incremental contribution.}
This work goes conservatively beyond mentioned prior works, by the conjunction of: (i) a set-up which disentangles surface position of a token from its potential information about future tokens and its true causal role in the data-generating process; (ii) considering a data-generating process with common causes, which are sometimes overt, sometimes latent; (iii) considering the emergence and mechanistic realization of ``Bayesian generalization'' for inferring a latent common cause in novel test cases.

\section{Experimental design, research questions \& approach}

The process for generating training and test data consists of repeatedly sampling triplets of token symbols and appending them into a long sequence, separated by a delimiter symbol, e.g., $aCb | feD | Zxy \dots$.
Each triplet of symbols is obtained by sampling a random \textit{context}, and then sampling that sequence exclusively from tokens belonging that context.
By designating particular contexts as \emph{held-out contexts} and withholding particular \textit{held-out sequences} from these held-out contexts during training, we are able to investigate cross-context generalization.

Fix $\mathcal{C} = \set{C_c}_{0 \le c \le n}$ as a family of \textit{contexts}, where each $C_c = \tuple{\mathcal{V}_c, G_c, P_c, A_{c}, M_{c}}$ consists of a causal Bayes net with three binary variables $\mathcal{V}_c = \set{\parent{c}{}, \child{c}{1}{}, \child{c}{2}{}}$ taking values in $\set{+, -}$, a causal dependency graph $G_c$ in which \parent{c}{} is a common cause of $\child{c}{1}{}$ and $\child{c}{2}{}$ ($\child{c}{1}{} \leftarrow \parent{c}{} \rightarrow \child{c}{2}{}$), so that the joint distribution $P_c$ is given by the causally disentangled factorization (in simplified notation):
\begin{align*}
  P_c\left(x, y^1, y^2\right) = P_c\left(x\right) \times P_c\left(y^1 \mid x\right) \times P_c\left(y^2 \mid x\right) \,,
\end{align*}
with $P_c(x) = \tfrac{1}{2}$ and $P_c(y^i \mid x) = p_c$ if $y^i = x$ and $1 - p_c$ otherwise.
Across contexts, Bayes nets share the variables, their values and the causal graph $G_c$, but they differ in $p_c$.
Moreover, $A_{c}$ is an alphabet (a set of six token symbols), so that $A_{c} \cap A_{c'} = \emptyset$, for all $c \neq c'$.
The mapping function $M_{c} \colon \mathcal{V}_{c} \times \set{+,-} \rightarrow A_{c}$ is a bijection, thus associating a unique symbol with each variable and value in $C_{c}$.
We call \parent{c}{} the \textit{parent}, and we call \child{c}{1}{} and \child{c}{2}{} the \textit{children}.
We say that \child{c}{i}{} is a \textit{sibling} to \child{c}{j}{} (implicitly assuming $i \neq j$ here and below).
The sampling protocol for one triplet sampled from context $C_{c}$ is: 
(i) sample a set of values from $P_c \left( \parent{c}{},\child{c}{1}{},\child{c}{2}{} \right)$;
(ii) map these values via $M_{c}$ to the corresponding token symbols in $A_{c}$;
(iii) randomly shuffle the three tokens symbols.

For readability, we represent (sets of) sequences of token symbols not in terms of elements of $A_{c}$ (as the language models will see them), but in terms of the corresponding variable-value pairs, writing $\child{c}{i}{-} \, \parent{c}{+} \,\, \child{c}{j}{+}$ or $\child{c}{i}{-} \, \child{c}{j}{+}$, where superscripts indicate the instantiated values, e.g., $\parent{c}{+} = M_{c}(\parent{c}{}, +)$.
We use the symbol $* \in \set{+,-}$ as a variable over values, and use $\bar{*} \in \set{+,-} \setminus \set{*}$ as its negation.
When we leave out subscripts or superscripts, we use this as notation for sets of sequences.
E.g., $\child{c}{i}{} \,\, \child{c}{j}{}$ denotes the set of all sequences of pairs of tokens, each of which is associated with a different child variable in context $c$, and notation like $\set{\parent{c}{}, \child{c}{}{}}$ would be shorthand for context $c$'s alphabet.
We say that a sequence is \textit{value-concordant} if all tokens in it are associated with the same value ($+$ or $-$); otherwise we speak of a \textit{value-discordant} sequence.
For instance, the set of all value-discordant sequences of child tokens from context $c$ would be written compactly as $\child{c}{i}{*} \, \child{c}{j}{\bar{*}}$.
We consider the context $C_{0}$ the \textit{held-out context}; all other contexts are referred to as \textit{supervised contexts}.
For all the \textit{held-out sequences} $\child{c_{0}}{i}{} \, \child{c_{0}}{j}{}$, the training data does not contain the succeeding parent token $\parent{c_{0}}{}$, and we refer to all other sequences as \textit{supervised sequences}. 


Our \textbf{main research question} is:
How could LMs learn a context-general form of (abductive) Bayesian reasoning from effect to cause by extrapolating the functional form of such reasoning observed from observed to unobserved cases? ---
We formulate a series of more specific research questions based on the assumption that sufficiently trained transformers will have found an approximately optimal solution to the training problem.
This perspective is kin to several approaches from different fields, e.g., \textit{rational analysis} in cognitive science \citep[e.g.,][]{Anderson1990:The-Adaptive-Ch}, the information-bottleneck approach in information theory \citep{TishbyPereira2000:The-information}, or computational mechanics in physics \citep[e.g.,][]{shaliziComputationalMechanicsPattern2001}.
But, going beyond a simple ideal-learner analysis, we additionally stress the constraints imposed the by the architectural realization of a computation, which is important in the context of generalization, as argued in the following.

An approach based on computational mechanics has recently been applied to the analysis of transformers' internal representations for reasoning under uncertainty as well \citep{shai2024transformers,levinson2026finding}.
To do so, computational mechanics identifies the \textit{canonical states} of the data-generating process, which are defined as the sufficient statistics for an optimal predictor, which, in turn, are given by the equivalence class of all sets of sequences that are indistinguishable based on the true probabilities of all future tokens \citep[cf.,][for similar methods]{VafaChen2024:Evaluating-the-}.
In our setup, all stochastic dependencies are broken by the delimiters and contexts have pairwise disjoint symbols, so that the canonical states are a union of the canonical states for each context $c$.
As, by construction, the only future-relevant aspect within a triplet for context $c$ is the value of the parent variable $X_c$, the canonical states are defined by tracking beliefs about $X_c$.
Consequently, we can characterize the canonical states via $\mathcal{S} = (\Omega, \Lambda)$ and $\Omega \subseteq \set{\parent{c}{}, \child{c}{}{}}$, 
where $\Omega$ is the set of variables already observed in the current triplet and $\Lambda$ is the true log-odds of the value of $X_c$.
We then only need to distinguish, based on the number of observed tokens so far, the following relevant canonical states:
\begin{align*}
  |\Omega| = 0: \ (\emptyset,\; 0) \quad \quad \quad
  |\Omega| = 1 &: \ (\{\parent{c}{}\},\; \pm\infty), \quad (\{\child{c}{}{}\},\; \pm\lambda_c) \\ 
  |\Omega| = 2 &: \ (\{\parent{c}{},\child{c}{}{}\},\; \pm\infty), \quad
                  (\{\child{c}{1}{},\child{c}{2}{}\},\; \{\pm 2\lambda_c,\, 0\})
\end{align*}
where $\lambda_c = \log \frac{p_c}{1-p_c}$ is the belief unit for context $c$.
By an ideal-learner analysis, we expect trained transformers to recover these canonical states, i.e., to track incrementally the log-odds of $X_c$ as the only prediction-critical information.
Evidence for this Bayesian-like ideal-learner behavior would come from high accuracy on training sequences, patterns of input-variance and -invariance that is consistent with the canonical states, part of which is the recovery of the patterns of stochastic independence between variables.
Moreover, we expect the ideal-learner analysis to be supported in the models' representations and their internal computational mechanisms.

What the perspective of computational mechanics and a pure ideal-learner analysis does not give us, is a prediction about generalization from training to held-out sequences.
The held-out sequences in context $C_{0}$ coincide with three canonical states $(\{\child{c}{1}{},\child{c}{2}{}\},\; \{\pm 2\lambda_c,\, 0\})$, i.e., with the four value-concordant two-child sequences $\child{c_{0}}{i}{*} \, \, \child{c_{0}}{j}{*}$ and the two value-discordant sequences $\child{c_{0}}{i}{*} \, \, \child{c_{0}}{j}{\bar{*}}$.
As all of these states are never seen in training, and ideal-learner perspective in terms of canonical state recovery does not predict whether or how generalization to these unseen cases may happen.
Instead, we hypothesize, based on architectural features, that transformers' hidden representations may capture similarities between canonical states in such a way that Bayes-like generalization is possible \citep{Shepard1987:Towards-a-Unive}.
Concretely, we conjecture that such generalization can occur if models capture (i) the mechanism of accumulating evidence (about the value regime) in a way that generalizes across (supervised) contexts, together with (ii) cross-context similarity of the evidential role of tokens/variables.
We therefore investigate whether such a similarity-based representation of token roles with a cross-contextually shared Bayesian-like mechanism of tracking of prediction-relevant evidence is attested in our trained models.

\section{Models, Training \& Behavioral Generalization}

We train small GPT-style decoder-only transformers over sequences comprised of $3$ contexts with $p_c=\{0.7, 0.8, 0.9\}$. $45$ models are trained in total (3 held-out contexts × 15 seeds; see Appendix~\ref{app:training} for full information). 
Each model has two pre-norm blocks with one attention head. 
Residual stream and feedforward layers (with GELU) both have a width of $16$, and each model a total of $4,051$ parameters. 
Token embeddings are learned and untied from the unembedding. 
Sinusoidal positional embeddings are used. 
Training uses next-token cross-entropy on one-hot targets, with AdamW and a fresh batch at every step. 
In the held-out context, the parent is removed from the loss and from the attention keys whenever it follows both children. 
We keep the checkpoint of each model with the lowest validation (never seen during training) KL-divergence to the exact Bayesian conditionals.
All trained models are used in all of the following analyses.

\begin{figure}[t]
  \centering
  \includegraphics[width=\linewidth]{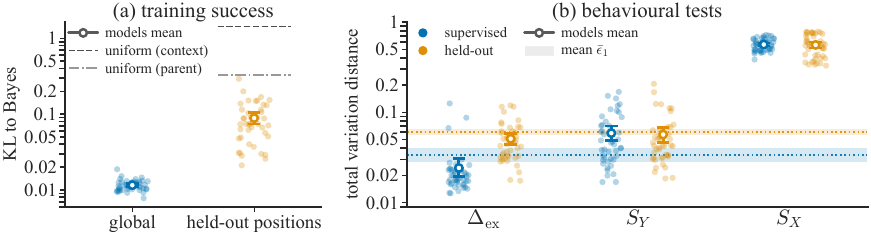}
  \caption{\textbf{Training success and stochastic dependencies in models' behavior.} 
  \textbf{(a)} Training success measures; \emph{(global)} averaged over all positions, and \emph{(held-out)} averaged over the masked positions in \textit{held-out sequences}. 
  \textbf{(b)} Total variation distances at second token of supervised~(blue) and held-out sequences~(orange). The three columns represent variable exchange error $\Delta_{\mathrm{ex}}$ (Eq.~\ref{eq:delta_ex}), child's predictive sensitivity to sibling $S_Y$ (Eq.~\ref{eq:S_Y}) and child's predictive sensitivity to parent $S_X$ (Eq.~\ref{eq:S_X}).
  Open markers: mean with 95\% two-level bootstrapped interval. 
  Shaded bands: $\bar{\epsilon}_1$ (Eq.~\ref{eq:predictive-error}), model's predictive error at the same position.
  $\Delta_{\mathrm{ex}}$ and $S_Y$, are in the range of models' error margin. $S_X$ reaches $0.56$, ground-truth is $|2p_c-1|$ (mean $0.60$): an order of magnitude above the floor.
  }
  \label{fig:behavior-stoch-dep}
\end{figure}

Figure~\ref{fig:behavior-stoch-dep}a shows that trained models achieve high predictive accuracy on the training data.
More importantly, although the held-out KL is an order of magnitude higher than the global KL-Divergence, it remains far below that of a random baseline. 
The trained models are particularly successful at predicting the correct variable at a held-out position. 
An optimal learner analysis predicts that for high predictive accuracy the patterns of stochastic dependencies entailed in the true causal graph must be recovered.
Figure~\ref{fig:behavior-stoch-dep}b shows that this is indeed so.
In both held-out and supervised contexts, models respect the exchangeability of variables and parents screen off children.
This suggests that successfully trained models recover general information about the \textit{role} of different tokens in line with the underlying (causal) variables.

\section{Representation of the data-generating causal model}
\label{sec:representation}

The behavioral results from above 
motivate subsequent questions about emergent representations.
\textbf{RQ1:} Do transformers trained on next-token prediction internalize concepts of variables and their values as specified by the data-generating process? 
\textbf{RQ2:} Do transformers carry a latent belief about the common cause? ---
To address these questions, Section~\ref{subsec:embeddings} shows that token embeddings they encode task-relevant information in a geometry that reflects the functional role of the variables in the true data-generating (causal) model.
Subsequently, Section~\ref{subsec:residual} shows that the functional attention combines these embeddings such that a belief coordinate carrying the direction of the evidence about the common cause emerges.
Finally, Section~\ref{subsec:accumulation-gain} shows that the model counts the number of observed variables utilizing the final LayerNorm.
These results are visualized in Figure~\ref{fig:results-overview}.

\subsection{Variable roles emerge in token embeddings}
\label{subsec:embeddings}

\begin{figure}[t]
    \centering
    \includegraphics[width=\linewidth]{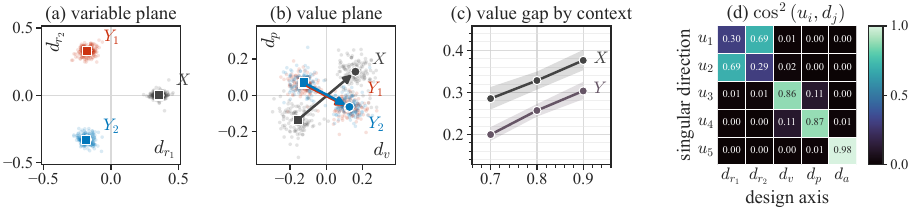}
    \caption{\textbf{Context-centered embeddings exhibit a common geometry.}
    \textbf{(a)} Variable plane encodes the variation among variable identities; $d_{r_1}$ separates parent from children, $d_{r_2}$ children from each other.
    \textbf{(b)} Value plane separates tokens by the regime value; $d_v$ and $d_r$ mostly captures a common value shift among all variables and parent's deviation from children, respectively.
    \textbf{(c)} Child values are exchangeable; their distance in the value plane tracks contexts dependency power (i.e., furthest for $p_c=0.9$, and shortest for $p_c=0.7$).
    \textbf{(d)} The design axes align with the natural directions that explain the variance of the raw token embeddings.}
    \label{fig:embeddings}
\end{figure}

We ask how each context's token embeddings represent that context's variables and their values, using an orthogonal decomposition of the embedding vectors (Appendix~\ref{app:embedding-decomposition}).
Let $\tilde{e}_c(V,*)$ denote the centered embedding of token $M_c(V,*)$ for all variables $V \in \mathcal{V}_c$ and values $* \in \set{+,-}$. Then define:
\begin{equation}
\tilde{e}_c(V,*) = \alpha_c(V) + \beta_c(*) + \gamma_c(V,*) \, ,
\end{equation}
where $\alpha_c(V)$ is the \textit{variable term}, $\beta_c(*)$ is the \textit{value term}, and $\gamma_c$ is the \textit{interaction term}.
Since the children are exchangeable, the variable and interaction planes each split further into a parent-vs-children axis and a child-vs-child axis.
To tie these design axes to the embedding geometry, we project the left singular vectors of the centered table (stack of token embeddings minus their context means) onto them and measure the fraction of each vector along each axis.

Each singular vector lies mostly along a single design axis, so the design axes are the principal axes of the embedding (Figure~\ref{fig:embeddings}d).
Among these axes, the variable term dominates, and the value and interaction terms are relatively small.
Almost all of the interaction plane is explained by the parent-vs-children axis.
The values of each context's variables are therefore represented in a two-dimensional plane, spanned by a common value direction and a parent-vs-children direction (Figure~\ref{fig:embeddings}b).
The two value tokens of a variable lie further apart in contexts with a stronger dependency, i.e., larger $p_c$ (Figure~\ref{fig:embeddings}c).
Taken together, these results answer \textbf{RQ1} at the level of token embeddings in two parts:
(1) \textbf{Token embeddings recover the variables and values of the generative process.} 
From next token prediction, the model represents each token as a variable (Figure~\ref{fig:embeddings}a) carrying a value (Figure~\ref{fig:embeddings}b), akin to the generative process. 
(2) \textbf{The embedding geometry reflects the structure of the dependency graph.}
The parent is distinguished from the two children by its value, while the children are exchangeable (Figure~\ref{fig:embeddings}b).

\subsection{A belief coordinate emerges after attention}
\label{subsec:residual}
Of the two attention layers in each model, only one is functional, and its pattern is fixed by position (Appendix~\ref{app:attn-loc}).
Such a content-invariant attention cannot itself distinguish variables or values.
Since the token embeddings already carry this distinction (Section~\ref{subsec:embeddings}), what remains is how the functional attention combines them.
Attention preserves the design decomposition of the embeddings up to a nearly uniform per-token LayerNorm scale (see Appendix~\ref{par:design-preservation}).
Since each model has a single functional attention block, there is a single residual node at which the observed tokens are combined as a weighted sum of their embeddings.
This simple attention can still function as part of a circuit that implements a posterior computation.
The missing piece is whether the token embeddings supply a form of \emph{evidence} that the attention can combine for prediction.

We study the residual stream of the trained models at the node right after a functional attention block, where the input is a two-token sequence $M_c(V,*)\,M_c(V',*')$ of distinct variables, $V \neq V' \in \mathcal{V}_c$.
We call the first token \emph{antecedent} and the second \emph{query} token.
We regress the average residual stream at the selected node, $\bar{h}_c$, on a global offset and two token terms over the $24$ admissible token pairs:
\begin{equation}
     \bar{h}_c(V,*;\,V',*') \;=\; o_c + a_c(V',*') + b_c(V,*) + \varepsilon_c(V,*;\,V',*') \, .
    \label{eq:rs-reg}
\end{equation}
For all $45$ trained transformers, the cross-term $\varepsilon_c$ is negligible, and the two token terms produce almost equal norms, $\|a_c\| \approx \|b_c\|$.
This suggests that each observation contributes separate evidence with the same weight as the other, regardless of the emission order.
To fix the scale of these terms, we take the residual stream of single-token sequences at the same residual node, $g_c(V,*)$, as reference.
We then measure the weight of each token term as $\|a_c\|/\|g_c\|$ and $\|b_c\|/\|g_c\|$.
In every context, both weights lie between $0.54$ and $0.57$ on average.
This establishes the \emph{functional} form of the attention block as an approximate average of the two tokens (See Appendix~\ref{app:functional-form} for more information).

Whether the geometries of the two token terms are aligned is a separate question.
We answer it with a principal-angle analysis of their design subspaces (See Appendix~\ref{app:functional-form}).
The principal angles are small for the value axis, both variable plane axes, and the leading interaction direction; only the second interaction direction, which carries little mass, is not aligned.
The attention block therefore averages the two tokens within each design subspace as well.

\begin{figure}[t]
    \centering
    \includegraphics[width=\linewidth]{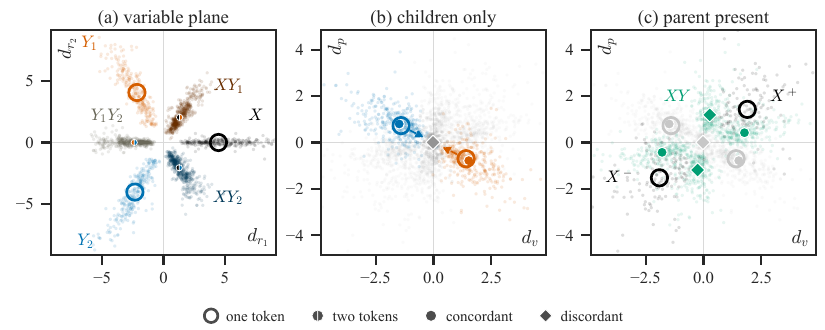}
    \caption{\textbf{Geometry of the residual stream.}
      Residual streams after functional attention, pooled over 45 models and all contexts, projected onto the
      design planes. Dots are pooled cells; markers are pool medians.
      \textbf{(a)}~Each variable and each pair occupies its own ray, a pair lying between its members.
      \textbf{(b)}~Value plane; in the absence of parent (gray dots), belief about it lies on the anti-diagonal. A concordant second
      child leaves the median in place and a discordant one returns it to
      the origin. Belief encodes the sign of the evidence and is insensitive to its count.
      \textbf{(c)}~Value plane (complementary); sequences containing the parent lie off the belief axis.}
    \label{fig:residual-geometry}
\end{figure}

This alignment has interesting implications.
First, the three centered variable vectors sum to zero, so the average of any two equals minus one half of the third.
The residual of a two-token sequence therefore always points away from the variable still to be emitted (Figure~\ref{fig:residual-geometry}a).
This alignment is almost exact in both supervised and held-out contexts, which explains why the models identify the variable of the held-out posterior correctly.
Second, the children move the residual stream along the same direction in value plane (Figure~\ref{fig:residual-geometry}b).
Value-concordant sequences shift it toward their shared value and value-discordant sequences cancel out.
This direction operates as a candidate \emph{belief coordinate}, which points at the value regime, i.e., whether current sequence favors one value of the parent over the other.
In other words, although the ground-truth belief takes five distinct values, the belief after attention is only ternary: it takes a positive or negative value for the sign of the evidence, and zero when the children disagree.

\subsection{Model counts the evidence}
\label{subsec:accumulation-gain}
A belief coordinate that only carries the direction of the observed evidence cannot by itself account for two-child sequences that produce a posterior of $\pm 2\lambda_c$. 
Additionally, the held-out sequences are indistinguishable from the supervised ones in the value plane.
Therefore, something after the attention has to amplify the belief in case of concordant two-child sequences such that the final read-out matches the ground-truth.
We hypothesize that the same mechanism accounts for the prediction deficit of the held-out posterior.

To locate this amplification, we project the output of each component downstream of the attention onto the parent's log-odds direction $w_c = \mathbf{u}(X_c^{+}) - \mathbf{u}(X_c^{-})$, the difference between the unembedding vectors of the two parent tokens of context $c$.
Since the residual stream is the sum of the components' outputs, its projection onto this direction is the sum of each component's output projection.
Before the output, the final LayerNorm standardizes the residual $h$ by subtracting its mean $\mu(h)$ and dividing by its scale $\sigma(h)$, and applies a gain $\gamma$ and a bias $b$. We can write the parent's log-odds as
\begin{equation*}
  v_c(h) = \frac{\big\langle \gamma \odot (h - \mu(h)\mathbf{1}),\, w_c \big\rangle}{\sigma(h)} + \langle b, w_c \rangle .
\end{equation*}
We then compare the residual of a single-child sequence $x^{(1)}$ with that of a value-concordant two-child sequence $x^{(2)}$.
The change in the parent's log-odds between them is
\begin{equation*}
  \frac{v_c(h^{(2)}) - \langle b, w_c \rangle}{v_c(h^{(1)}) - \langle b, w_c \rangle}
  = \frac{\big\langle \gamma \odot (h^{(2)} - \mu(h^{(2)})\mathbf{1}),\, w_c \big\rangle}{\big\langle \gamma \odot (h^{(1)} - \mu(h^{(1)})\mathbf{1}),\, w_c \big\rangle}
  \cdot \frac{\sigma(h^{(1)})}{\sigma(h^{(2)})} .
\end{equation*}
where the first factor is carried by the model's other components and the second by the final LayerNorm.

\begin{figure}[t]
    \centering
    \includegraphics[width=\linewidth]{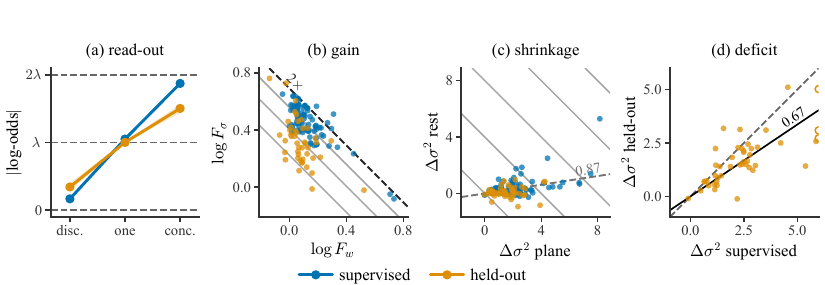}
    \caption{\textbf{Accumulation gain.}
    \textbf{(a)} Bootstrap mean with 95\% CI of absolute parent log-odds in units of $|\lambda_c|$. Held-out two-child cases are not exact.
    \textbf{(b)} Amplification of the parent's log-odds from one child to two concordant children, split into the factor from the components, $F_w$, and from the final LayerNorm, $F_\sigma$. Grey diagonals mark total gains of $1.25$, $1.5$ and $1.75\times$; the dashed diagonal marks $2\times$.
    \textbf{(c)} Shrinkage of the final LayerNorm input from one child to two concordant children, $\Delta\sigma^2$, split into the part inside the variable plane and the rest. Grey diagonals mark constant total shrinkage; the dashed line marks the pooled in-plane share, $\sum \Delta\sigma^2_{\mathrm{plane}} / \sum \Delta\sigma^2$.
    \textbf{(d)} $\Delta\sigma^2$ in the held-out context against the mean over the two supervised contexts, one point per model. Dashed: identity; solid: pooled ratio $\sum \Delta\sigma^2_{\text{held-out}} / \sum \Delta\sigma^2_{\text{supervised}}$; open markers are clipped at the axis limit.}
    \label{fig:readout-gain}
\end{figure}

Normalization accounts for most of the amplification (Figure~\ref{fig:readout-gain}), which it achieves because the residual shrinks from a one-child sequence to two.
Splitting this shrinkage into the part inside the variable plane and the part outside shows the plane carries most of the shrinkage.
The amplification is therefore set by the count of observed variables.
We call this mechanism the \emph{accumulation gain}.
In the held-out context, the one-child log-odds match $\lambda_c$ but the two-child log-odds fall short of $2\lambda_c$ (Figure~\ref{fig:readout-gain}a). 
This is explained by the residual shrinkage of the held-out states, which is only $67\%$ of that in the supervised contexts (See Appendix~\ref{app:plane-shrinkage}).
Smaller shrinkage means a larger activation magnitude, hence a smaller normalization factor and a mismatch in prediction.
Next, we ask what governs this mismatch in shrinkage.
Expanding the normalization term $\sigma^2(h)$ over the model's components shows how much each pair of components align with each other, and hence how much they lengthen or shorten the residual (See Appendix~\ref{app:sigma-expansion}).
The expansion shows that the held-out deficit results from multiple small misalignments between components.
This explains why the approximate geometric analyses could not resolve them.

In sum, we resolve \textbf{RQ2} with two results.
(1) \textbf{The residual stream encodes a belief about the common cause.}
    The shared value direction is represented as a belief coordinate after the attention which encodes \emph{only} the direction of the evidence about the parent, hence a latent belief.
(2) \textbf{The model counts the evidence toward the common cause.}
    The unit evidence direction in the belief coordinate is multiplied by the number of observed child tokens, via the accumulation gain.

\section{Joint Dependency Intervention}
Thus far, we have established that the trained transformers internalize the structure of the data-generating process and encode a latent belief.
We argue that these are only prerequisites for ascribing the adjective Bayesian to the transformers' computation.
Building on the causal abstraction framework \cite{GeigerLu2021:Causal-Abstract}, Bayesian inference can only be a faithful causal abstraction of that computation if the represented latent \emph{operates} as a Bayesian latent.
A Bayesian latent about the common cause under the fork structure of $G_c$ determines the \emph{joint distribution} over the unobserved variables given one observed child variable token.
This means that, given an intervention on the represented latent, both the posterior over the parent and the prediction of the sibling should change consistently with Bayes's rule.
Therefore we pose \textbf{RQ3}: Does the represented latent determine the joint distribution over the unobserved variables as prescribed by Bayesian inference?

Based on results from Section~\ref{sec:representation}, we can describe outputs in log-odds coordinates as:
\begin{equation}
    \hat\Lambda_{\parent{c}{}} = G(|\Omega|)\, \rho_{\parent{c}{}}\, z, \qquad \hat\Lambda_{\child{c}{j}{}} = \rho_{\child{c}{j}{}}\, z,
    \label{eq:functional-form}
\end{equation}
where $z \in \{0, \pm 1\}$ is the belief coordinate in unit beliefs, $G(|\Omega|) = \sigma(1)/\sigma(|\Omega|)$ is the accumulation gain, with $\sigma(|\Omega|)$ the scale of the final normalization when $|\Omega|$ children are observed, and $\rho_{\parent{c}{}}$ and $\rho_{\child{c}{j}{}}$ are the context-specific readout factors mapping the belief coordinate onto the log-odds of the parent and of the sibling.
We use Eq.~(\ref{eq:functional-form}) to fix the alignment between the transformer's states and the variables of the Bayesian model.
We then test this by a \emph{joint dependency intervention}.
We intervene on the represented latent at the residual node after the functional attention, and add $n$ unit beliefs along $z$.
We compare the resulting shifts in the log-odds of the parent and the sibling with those prescribed by Bayes's rule.
We read the parent at one- and two-child sequences, and the sibling at one-child sequences.
Bayes's rule couples the log-odds of the sibling to those of the parent, and we linearize this coupling by passing to $\tanh$ coordinates:
\begin{equation}
    \tanh\!\left(\hat\Lambda_{\child{c}{j}{}}/2\right) = (2p_c - 1)\tanh\!\left(\hat\Lambda_{\parent{c}{}}/2\right).
\end{equation}
\begin{figure}
    \centering
    \includegraphics[width=\linewidth]{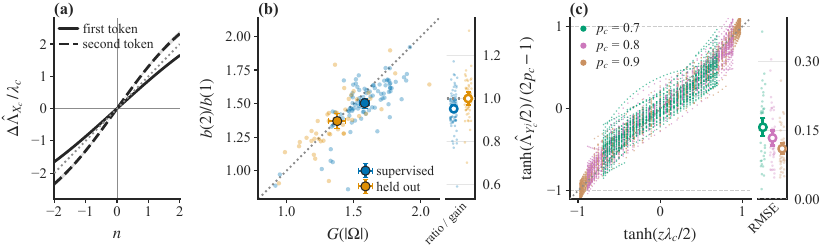}
    \caption{\textbf{Joint dependency intervention.}
    $n$ unit beliefs are added along the belief coordinate after the functional attention; 135 (model, context) cells, means with $95\%$ bootstrap CIs.
    \textbf{(a)} Parent log-odds shift against $n$ at the first (solid) and second (dashed) child positions.
    \textbf{(b)} Slope of the two-child over the one-child line from (a) as \emph
    {intervention ratio} against the \emph{accumulation gain} $G(|\Omega|)$, supervised and held-out cluster; right: intervention ratio over accumulation gain is close to one, i.e., intervention follows the functional form.
    \textbf{(c)} Sibling against parent in $\tanh$ coordinates, colored by context index $p_c$; right: residual mean squared error from the Bayes predicted identity line.}
    \label{fig:jdi}
\end{figure}

At one-child sequences, one injected unit belief shifts the parent's log-odds by $\rho_{\parent{c}{}}$, slightly below $\lambda_c$ (Figure~\ref{fig:jdi}a).
At two-child sequences, the shift is larger by the accumulation gain $G(2)$ (Figure~\ref{fig:jdi}b).
In $\tanh$ coordinates, the sibling follows the identity line across the steering range (Figure~\ref{fig:jdi}c), with a with the minimum error when the dependency is strongest.
Steering the represented latent therefore shifts the prediction of the unobserved sibling by the amount Bayes's rule prescribes.
The joint dependency intervention thereby answers \textbf{RQ3} by showting that \textbf{the represented latent operates as a Bayesian latent}, i.e., changing it along the belief coordinate shifts the joint distribution over the unobserved variables consistently with Bayes's rule.

\section{Discussion, Conclusions, Limitations \& Outlook}

We presented a case study of transformers learning generalizable Bayesian abductive reasoning.
Conceptually, we separated within-model from behind-data causality, and adopted an architecturally-aware optimal learner perspective to structure analysis.
Methodologically, we designed a minimal setup that partially disentangles the two notions of causality.
Empirically, we showed, supported by a joint dependency intervention analysis, that ``Bayesian generalization'' rests on representations shared across contexts, a categorical belief coordinate and a mapping to Bayesian posteriors (see Figure~\ref{fig:results-overview}).
%
%
The uncovered mechanism is an implementation of genuine Bayesian inference for the task. 
However, in our setup, it is not possible to recover the precise causal role of the parent variable due to known theoretical limits of \textit{causal recovery} \citep{spirtes2000causation,richardson2002ancestral}.
Our results do not transfer directly to large models, but suggest that they may also achieve ``Bayesian generalization'' by exploiting embedding similarity and efficient approximations to Bayesian computation.

Future work could continue the work presented here, e.g., by looking at more substantial (hierarchical) latent generation models, where latent variables are never observed.
Models trained to predict tokens that are themselves informative about the data-generating process of other tokens might also make it possible to study ``represented causality'' in controlled experimental settings. 

\newpage 

\section{Statements}
\subsection*{AI use statement}

In this work, we used generative AI tools for coding the experiments, implementing analyses and visualizations, drafting parts of the appendices, and feedback on the experimental design and the interpretation of results. We did not use generative AI tools for developing the conceptual framework, data generation, mathematical proofs, translation, or writing the main text. We have reviewed all AI-assisted work and take full responsibility for the final content, including any text or artifacts produced with the aid of generative AI.

\subsubsection*{Acknowledgments}

 AM and MF were supported by the Volkswagen Foundation through a Momentum grant.
 MF is a member of the Machine Learning Cluster of Excellence at University of T\"ubingen, EXC number 2064/2 – Project number 39072764. 
 AM and MF gratefully acknowledge support by the state of Baden- W\"urttemberg through bwHPC and the German Research Foundation (DFG) through grant INST 35/1597-1 FUGG.
 
\bibliography{custom}
\bibliographystyle{iclr2027_conference}

\appendix
\clearpage
\section{Training}
\label{app:training}

\paragraph{Architecture.}
GPT-style pre-LayerNorm decoder-only transformers with untied embedding and unembedding (Table~\ref{tab:arch}).
\begin{table}[h]
\centering
\caption{Architecture configuration common to all 45 models.}
\label{tab:arch}
\begin{tabular}{ll}
\toprule
Layers / heads per layer & 2 / 1 \\
$d_{\text{model}}$ / head dimension & 16 / 16 \\
MLP hidden size, activation & 16, GELU \\
Dropout & none \\
Biases & QKV and output projections \\
Positional encoding & sinusoidal, max length 1000 \\
Vocabulary & 19 (1 shared delimiter $+$ 3 contexts $\times$ 6) \\
Trainable parameters & 4{,}051 \\
\bottomrule
\end{tabular}
\end{table}
\paragraph{Data sampling.}
Each sequence is sampled by some stratification constraints from the data-generating process to provide a closer match to theoretical distribution. In a 60 triplet long sequence, each context is represented 20 times, and within each context, each value of $\parent{c}{}$ appears 10 times, and the value counts of $\child{c}{1}{}$ and $\child{c}{2}{}$ match $p_c$. The emission order is sampled uniformly (approximately if counts permit) per each context. There is no fixed training set. A fresh batch is drawn at every step.
\paragraph{Objective and held-out masking.}
The loss is cross-entropy against one-hot next-token targets over all content positions; positions whose target is the delimiter are excluded.
Each model has one held-out context, $C_{0}$.
In every triplet of $C_{0}$ ordered as $\child{c_0}{i}{}\,\child{c_0}{j}{}\,\parent{c_0}{}$, the parent token is removed from both the loss and the attention keys.
As a result, the model is never trained to predict the parent from a held-out sequence $\child{0}{i}{}\,\child{0}{j}{}$.
All other positions of $C_{0}$ are trained normally.
\paragraph{Training.}
Table~\ref{tab:training} lists the optimization and checkpoint-selection settings. A single
seed per model sets both the initialization and the data stream. Checkpoints are selected by
$\mathrm{KL}\!\left(P_{\text{Bayes}} \,\|\, P_\theta\right)$ on the validation set, where
$P_{\text{Bayes}}$ is the exact Bayesian conditional, averaged over contexts.

\begin{table}[H]
\centering
\caption{Optimization and checkpoint selection.}
\label{tab:training}
\begin{tabular}{ll}
\toprule
Optimizer & AdamW, $\beta = (0.9, 0.999)$, weight decay $0.1$ \\
Learning rate & $0.05$, constant (no warmup or decay) \\
Gradient clipping & norm $1.0$, no accumulation \\
Batch size & 384 sequences (92{,}160 tokens) \\
Training budget & 20k steps (7.68M sequences) \\
\midrule
Validation set & 4{,}800 fixed sequences \\
Evaluation / checkpoint interval & every 10 steps \\
Selection criterion & $\min \mathrm{KL}(P_{\text{Bayes}} \,\|\, P_\theta)$ \\
Selected step & 2{,}390--17{,}480, median 8{,}450 \\
Selected sequences seen & 0.9M--6.7M, median 3.2M \\
\bottomrule
\end{tabular}
\end{table}

\paragraph{Trained models.}
We train 45 models, crossing the held-out context parameter $p_0 \in \{0.7, 0.8, 0.9\}$ with 15
seeds $\{42, 100, 110, 120, 130, 140, 150, 200, 300, 400, 500, 600, 700, 800, 900\}$. Seeds repeat across held-out conditions, so each model is identified by its
$(p_0, \text{seed})$ pair. Table~\ref{tab:models} shows the emergence of a sparse attention layer. The sparse layer is the layer whose attention has the
largest mean $\mathrm{KL}(\text{attention row} \,\|\, \text{uniform})$, and the other layer is diffuse; Appendix~\ref{app:attn-loc} shows that the sparse layer is also the functional layer.

\begin{table}[H]
\centering
\caption{Sparse attention layer by held-out context.}
\label{tab:models}
\begin{tabular}{lcccc}
\toprule
SparseYour  attention & Total & $p_0 = 0.7$ & $p_0 = 0.8$ & $p_0 = 0.9$ \\
\midrule
Layer 1 & 31 & 8 & 13 & 10 \\
Layer 2 & 14 & 7 & 2 & 5 \\
\bottomrule
\end{tabular}
\end{table}

\section{Behavioral Metrics}
\label{app:behavior-metrics}
\paragraph{ Prediction invariance to emission order.}
The conditional future should not be a function of the permutation latent over the observed evidence.
We measure to what extent this is true by \emph{exchangeability error} $\Delta_{\text{ex}}$ derived by the total variation distance (TVD) of the model's distributions before and after swapping the two observed tokens in a two-token sequence
\begin{equation}
    \Delta_{\mathrm{ex}} = \delta\Big( P_\theta\big(\cdot \mid M_c(v,*)\,M_c(v',*')\big),\; P_\theta\big(\cdot \mid M_c(v',*')\,M_c(v,*)\big) \Big)
    \label{eq:delta_ex}
\end{equation}
\paragraph{The parent d-separates the children.}
The conditional future should only be a function of the parent's value.
Two prediction sensitivity measures quantify this claim in models, namely sensitivity of prediction of a child to its sibling $S_Y$, and to its parent $S_X$. They are both derived by the TVD of model's two distributions, before and after a value flip. We interpret them up to a predictive error of the model at the same token position as defined below.
\begin{align}
        S_Y &= \delta\Big( P_\theta\big(\cdot \mid \parent{c}{*}\child{c}{i}{*}\big),\; P_\theta\big(\cdot \mid \parent{c}{*}\child{c}{i}{\bar{*}}\big) \Big) \label{eq:S_Y}\\
        S_X &= \delta\Big( P_\theta\big(\cdot \mid \parent{c}{*}\child{c}{i}{*}\big),\; P_\theta\big(\cdot \mid \parent{c}{\bar{*}}\child{c}{i}{*}\big) \Big) \label{eq:S_X}\\
        \bar{\epsilon}_1 &= \delta\Big( P_\theta\big(\cdot \mid M_c(v,*)\,M_c(v',*')\big),\; P_c\big(\cdot \mid M_c(v,*)\,M_c(v',*')\big) \Big) \label{eq:predictive-error}
\end{align}

\section{Orthogonal decomposition of the embedding table}
\label{app:embedding-decomposition}

For each model and context $C_c$, the six embeddings $e_c(V,*) = e\big(M_c(V,*)\big) \in \mathbb{R}^{16}$, with $V \in \mathcal{V}_c$ and $* \in \set{+,-}$, form a $3 \times 2$ table of vectors. We stack them as the rows of $\tilde E_c \in \mathbb{R}^{6 \times 16}$ after subtracting their mean, in the order $(\parent{c}{},+), (\parent{c}{},-), (\child{c}{1}{},+), (\child{c}{1}{},-), (\child{c}{2}{},+), (\child{c}{2}{},-)$. The centered table has rank at most five.

The usual two-way decomposition splits the five-dimensional column space of $\tilde E_c$ into three subspaces namely, a two-dimensional variable plane, a one-dimensional value axis, and a two-dimensional interaction plane. Since the children are exchangeable under the generative process, we split each plane once more into a parent-vs-children and a child-vs-child part. This gives five orthonormal axes in $\mathbb{R}^6$, the variable axes $d_{r_1}, d_{r_2}$, the value axis $d_v$, and the interaction axes $d_p, d_a$,
\begin{align}
d_{r_1} &= \tfrac{1}{\sqrt{12}}(2,2,-1,-1,-1,-1), & d_{r_2} &= \tfrac12(0,0,1,1,-1,-1), \nonumber\\
d_{p} &= \tfrac{1}{\sqrt{12}}(2,-2,-1,1,-1,1), & d_{a} &= \tfrac12(0,0,1,-1,-1,1), \nonumber\\
d_{v} &= \tfrac{1}{\sqrt6}(1,-1,1,-1,1,-1). & &
\end{align}
$d_{r_1}$ separates the parent from the children and $d_{r_2}$ the two children from each other. $d_v$ is the value shift common to all three variables. $d_p$ captures how the parent's value shift departs from the common one, and $d_a$ how the two children's shifts differ. Orthogonality gives
\begin{equation}
\|\tilde E_c\|_F^2 = \sum_{k \in \{r_1, r_2, v, p, a\}} \big\|d_k^{\top} \tilde E_c\big\|^2 .
\label{eq:ss-identity}
\end{equation}
Finally, we ask whether these label-defined axes are also the principal axes of $\tilde E_c$. With $u_{c,j}$ the $j$-th left singular vector of $\tilde E_c$, its weight on axis $k$ is the squared cosine
\begin{equation}
\big(d_k^{\top} u_{c,j}\big)^2, \qquad \textstyle\sum_{k} \big(d_k^{\top} u_{c,j}\big)^2 = 1 .
\end{equation}
\begin{figure}
    \centering
    \includegraphics[width=\linewidth]{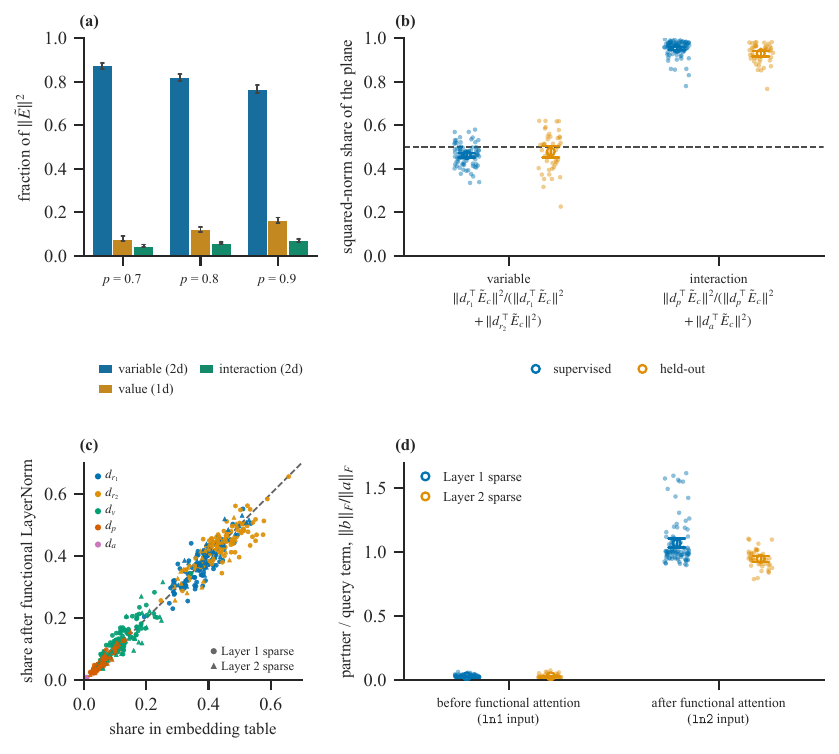}
    \caption{For each model and context $C_c$, the centered table $\tilde E_c \in \mathbb{R}^{6\times16}$, one row per variable and value, is split into the variable plane, the value axis and the interaction plane (Appendix~\ref{app:embedding-decomposition}).
    \textbf{(a)} Share of $\|\tilde E_c\|_F^2$ in each subspace, by context $p_c$. Bars are means over the 45 models, and error bars are 95\% bootstrap CIs with the model as resampling unit. This split uses no parent/child structure. Most of the table lies in the variable plane. The value share grows with $p_c$, and the interaction share stays small.
    \textbf{(b)} Share of the parent-vs-children axis within each plane, $\|d_{r_1}^\top \tilde E_c\|^2 / (\|d_{r_1}^\top \tilde E_c\|^2 + \|d_{r_2}^\top \tilde E_c\|^2)$ for the variable plane and $\|d_{p}^\top \tilde E_c\|^2 / (\|d_{p}^\top \tilde E_c\|^2 + \|d_{a}^\top \tilde E_c\|^2)$ for the interaction plane. Dots are single (model, context) values, and open markers are means with 95\% bootstrap CIs. The dashed line at $0.5$ marks a plane with no parent/child asymmetry. The variable plane lies near it, and the interaction plane is almost entirely the parent deviation. Supervised and held-out contexts agree throughout.
    \textbf{(c)} Share of each axis before and after the LayerNorm that precedes the functional attention, for the first, second and third token of a triplet. $d_{r_1}$ separates the parent from the children, $d_{r_2}$ the two children, $d_v$ is the common value shift, $d_p$ the parent's departure from it, and $d_a$ the difference between the children's shifts. Points lie on the identity line, so the LayerNorm leaves the decomposition intact.
    \textbf{(d)} Partner term relative to query term, $\|b\|_F/\|a\|_F$, at the second token of a triplet, where $a$ and $b$ are the query's and the partner's contributions to the cell means of the residual stream. Markers as in (b). Before the functional attention (\texttt{ln1} input) the partner term is essentially zero. After it (\texttt{ln2} input), the ratio is close to 1 in both model groups, so attention adds the partner's embedding to the query's own at equal weight.
    }
    \label{fig:embeddings-app}
\end{figure}

\paragraph{Preservation through the functional attention.}
\label{par:design-preservation}
The axes are defined on the token index, so they apply wherever the six tokens of a context can be tabulated. Additionally centering removes the positional information. The pre-attention LayerNorm standardizes and rescales each row separately, yet the shares $\|d_k^{\top} \tilde E_c\|^2 / \|\tilde E_c\|_F^2$ of its output match those of the embedding table (Figure~\ref{fig:embeddings-app}, so the decomposition enters the attention intact. The value-output map acts on the residual dimension while the axes act on the token index, hence
\begin{equation}
d_k^{\top}\big(\tilde E_c W_{OV}\big) = \big(d_k^{\top} \tilde E_c\big)\, W_{OV} \qquad \text{for every } k .
\label{eq:ov-commute}
\end{equation}
Attention thus carries each component through separately, changing only its norm and orientation.

\section{Attention Localization}
\label{app:attn-loc}
We isolate the within-model causal effects of each attention block by activation patching.
Let $h_{\text{base}}, h_{\text{donor}}$ be two sampled histories, each followed by a final triplet from context $C_c$.
The two final triplets agree on every token before the last one and differ only in the canonical state $\mathcal{S} = (\Omega, \Lambda)$.
We run two separate forward passes, the base and the donor run, and write $\mathrm{att}_L^{\text{base}}, \mathrm{att}_L^{\text{donor}}$ for the
layer-$L$ attention output at the last token.
The patched run is then inference on the base sequence with substituted activations from the donor, at
the last token and layer $L$.
We patch at the first ($|\Omega| = 1$) and the second ($|\Omega| = 2$) token of the final triplet.
We additionally draw $h_{\text{donor}}$ with an independent history from $C_c$, and with an
independent history from $C_{c'}$, $c' \neq c$.

Let $\ell_{\text{base}}, \ell_{\text{donor}}$ be the logits from the two clean runs and $\ell_{\text{patch}}$ the
logits from the patched run, each over the full vocabulary $\set{\text{delimiter}} \cup \bigcup_c A_c$. We define the recovery measure as
the fraction of the logit difference between the base and the donor run that the patch recovers,
\[
R_L \;=\;
\frac{\langle \ell_{\text{patch}}-\ell_{\text{base}},\;\ell_{\text{donor}}-\ell_{\text{base}}\rangle}
     {\langle \ell_{\text{donor}}-\ell_{\text{base}},\;\ell_{\text{donor}}-\ell_{\text{base}}\rangle}.
\]
$R_L=1$ means $\mathrm{att}_L$ carries the entire state distinction, $R_L=0$ means it carries none.

\begin{figure}[h]
    \begin{subfigure}{\linewidth}
        \centering
        \includegraphics[width=\linewidth]{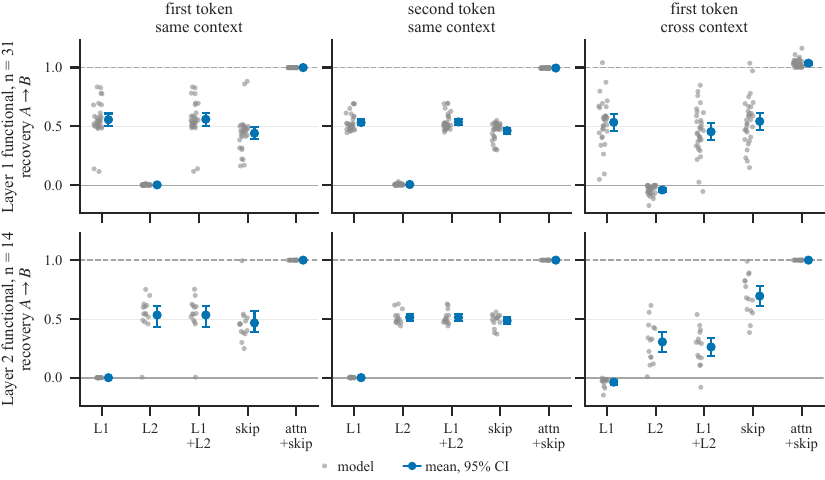}
        \caption{Caption}
        \label{fig:interchange}
    \end{subfigure}
    \\
    \begin{subfigure}{\linewidth}
        \centering
        \includegraphics[width=\linewidth]{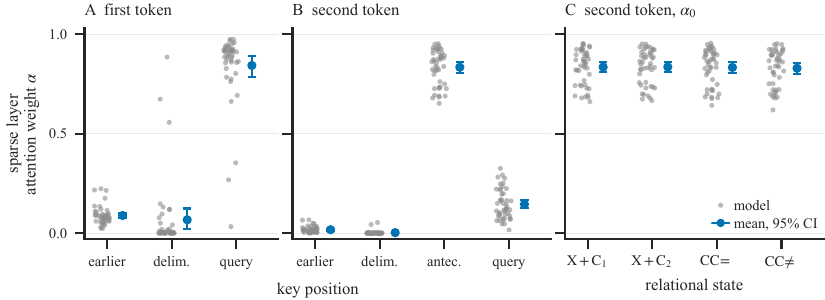}
        \caption{Caption}
        \label{fig:invariance}
        \end{subfigure}
\end{figure}

The models follow the categorization by attention patterns sparsity.
In $31$ models where the first layer is sparse, that layer recovers a large fraction of the distinction whereas the second layer recovers almost nothing. 
The reverse holds for the $14$ models with the second layer being sparse.
(Figure~\ref{fig:interchange}). Recovery is unchanged when the donor history is drawn independently from $C_c$.
Patching both attention layers together yields the same recovery as patching the sparse layer alone, therefore the diffuse attention carries no additional information.
The recovery of sparse attention is below one, then the skip connection must carry some of the distinction.
Patching the connection alone recovers $0.469$ of the distinction, therefore together they account for the entire distinction.

\section{Functional form of the attention block}
\label{app:functional-form}

We treat the skip connection and the functional attention as one block and read everything at the residual node that follows it, in unnormalized coordinates. Write $h_c$ for the residual stream at that node.

\paragraph{Regression.}
For context $C_c$, consider the second token of a triplet. The two tokens observed so far are $M_c(V,*)\,M_c(V',*')$, with $V \neq V' \in \mathcal{V}_c$ and $*, *' \in \set{+,-}$. Let $\bar h_c(V,*;\,V',*')$ be the mean of $h_c$ over all instances of a pair, that is, over all histories that precede it. We arrange these means as a $6 \times 6$ table of vectors in $\mathbb{R}^{16}$, with rows indexed by the query token $M_c(V',*')$ and columns by the antecedent token $M_c(V,*)$, both in the row order of $\tilde E_c$. Pairs with $V = V'$ do not occur, so the three diagonal $2 \times 2$ blocks are empty, and each row holds four cells, $24$ in total. Likewise, let $g_c(V,*)$ be the mean of $h_c$ at the first token of a triplet whose token is $M_c(V,*)$.

We model the table as a global offset plus a row term and a column term,
\begin{equation}
\bar h_c(V,*;\,V',*') \;=\; o_c + a_c(V',*') + b_c(V,*) + \varepsilon_c(V,*;\,V',*'),
\label{eq:app-rs-reg}
\end{equation}
where $a_c$ is the query term, $b_c$ the antecedent term, and $\varepsilon_c$ the cross-term. The terms are made unique by requiring that the means of $a_c$ and $b_c$, each weighted by how often its token occurs as query or antecedent, are zero; $o_c$ is then the frequency-weighted grand mean of the table.

Because the empty cells lie on the diagonal blocks, a row mean averages only over the antecedents of the two other variables. With balanced counts, the antecedent terms in a row sum to $-\big(b_c(V',+) + b_c(V',-)\big)$, so the row mean holds
\[
o_c + a_c(V',*') - \tfrac14\big(b_c(V',+) + b_c(V',-)\big),
\]
a remainder that depends on the query's variable. The same holds for column means. We therefore estimate all terms jointly by minimizing
\begin{equation}
\mathbb{E}\,\big\| h_c - o_c - a_c(V',*') - b_c(V,*) \big\|^2
\end{equation}
over all instances of two-token prefixes in the data. This equals the least-squares fit of the $24$ occupied cells, weighted by their frequencies.

\begin{figure}[t]
    \centering
    \includegraphics[width=\linewidth]{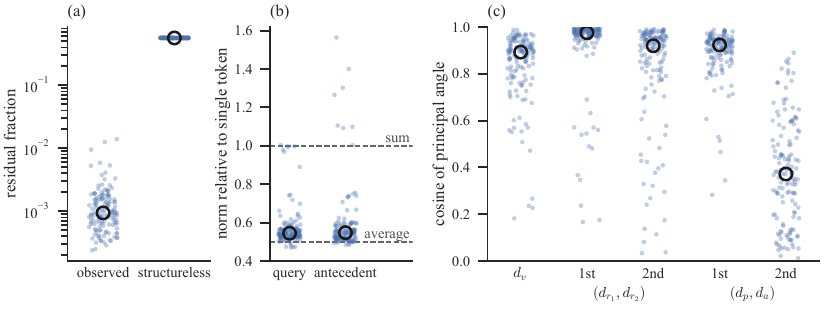}
    \caption{Functional form of the attention block, read at the residual node after the functional attention for the second token of a triplet. Each dot is one model and context; open circles are medians.
    \textbf{(a)} Share of the residual stream that a sum of one query term and one antecedent term cannot explain (log scale). The observed share is compared to that of random tables fitted in the same way.
    \textbf{(b)} Size of the query and antecedent terms relative to the residual of the same token observed alone. Dashed lines mark the values expected if the node summed the two tokens (1) or averaged them (0.5).
    \textbf{(c)} Alignment of the query and antecedent terms within each design subspace: the value axis $d_v$, the variable plane $(d_{r_1}, d_{r_2})$ and the interaction plane $(d_p, d_a)$. For each plane, 1st and 2nd are its two principal angles, from most to least aligned; a cosine of 1 means the two terms point in the same direction.}
    \label{fig:app-functional-form}
\end{figure}

\paragraph{Additivity.}
We measure the cross-term by the residual fraction
\begin{equation}
\frac{\mathbb{E}\,\|\varepsilon_c\|^2}{\mathbb{E}\,\|\bar h_c - o_c\|^2},
\end{equation}
with both expectations over pairs, weighted by their frequencies. On this layout, the fit has $11$ free parameters per residual dimension ($1$ for the offset and $5$ for each token term), so even a table without structure leaves a residual fraction bounded away from zero. For balanced counts, its expected value is $13/23$: the residual degrees of freedom over the centered ones. As a reference, we fit independent standard-normal tables with the same layout and frequencies in the same way. The observed residual fraction lies orders of magnitude below this reference (Figure~\ref{fig:app-functional-form}a), so nothing at the node depends on the pair beyond what each token contributes alone.

\paragraph{Fixing the scale.}
The regression leaves open whether the node sums or averages the two tokens. We take the single-token residual $g_c$ as reference. With $a_c$, $b_c$ and $g_c$ stacked as $6 \times 16$ tables in the row order of $\tilde E_c$, centered over their rows, and $\|\cdot\|$ the Frobenius norm, summation implies $\|a_c\| / \|g_c\| = \|b_c\| / \|g_c\| = 1$, and averaging implies $\tfrac12$. Both ratios lie near $\tfrac12$ in every context (Figure~\ref{fig:app-functional-form}b), so, up to the offset, the node holds approximately the average of the two single-token residuals.

\paragraph{Subspace alignment.}
Equal norms do not imply that the two terms occupy the same directions. For each design subspace, the value axis $d_v$, the variable plane $\mathrm{span}(d_{r_1}, d_{r_2})$ and the interaction plane $\mathrm{span}(d_p, d_a)$, we project $a_c$ and $b_c$ onto it. The projections of each term span a subspace of $\mathbb{R}^{16}$: one-dimensional for the value axis and two-dimensional for each plane. Between the two terms' subspaces, we compute the principal angles, the arccosines of the singular values of the product of their orthonormal bases. Principal angles depend on the subspaces and not on the bases chosen for them, so the result is the same for any basis of each design plane (Figure~\ref{fig:app-functional-form}c). The angle of the value axis, both principal angles of the variable plane, and the first principal angle of the interaction plane are small, in supervised and held-out contexts alike: their mean cosines differ between the two by at most $0.02$ on the value axis and the variable plane. Only the second interaction angle is large, in a direction that carries negligible mass.

\section{Normalization at the read-out}
\label{app:readout-normalization}

\paragraph{Read-out.}
Let $h \in \mathbb{R}^{d}$ be the residual entering the final LayerNorm, with $\mu(h) = \tfrac{1}{d}\mathbf{1}^{\top} h$ and $\sigma(h) = \big(\tfrac{1}{d}\lVert h - \mu(h)\mathbf{1} \rVert^{2} + \epsilon\big)^{1/2}$.
The LayerNorm maps $h$ to $\gamma \odot (h - \mu(h)\mathbf{1}) / \sigma(h) + b$.
With $w_c = \mathbf{u}(X_c^{+}) - \mathbf{u}(X_c^{-})$, the parent's log-odds are $v_c(h) = A(h) + \langle b, w_c \rangle$, with
\begin{equation}
  A(h) = \frac{N(h)}{\sigma(h)} ,
  \qquad
  N(h) = \big\langle \gamma \odot (h - \mu(h)\mathbf{1}),\, w_c \big\rangle .
\end{equation}

\paragraph{Gain.}
Let $h^{(1)}$ and $h^{(2)}$ be the residual stream activations before the final LayerNorm of a single-child and a value-concordant two-child sequence, respectively.
With $\mathbb{E}^{(1)}$ and $\mathbb{E}^{(2)}$ the means over the single-child and value-concordant two-child sequences of one model and context,
\begin{equation}
  F = \frac{\mathbb{E}^{(2)}[A]}{\mathbb{E}^{(1)}[A]} = F_w \, F_\sigma ,
  \qquad
  F_w = \frac{\mathbb{E}^{(2)}[N]}{\mathbb{E}^{(1)}[N]} ,
  \qquad
  \log F = \log F_w + \log F_\sigma .
\end{equation}

\paragraph{Writers.}
The residual $h$ is the sum of five writer outputs, $h = e + a_1 + f_1 + a_2 + f_2$, with $e$ the token and position embeddings and $a_\ell$ and $f_\ell$ the attention and MLP (feed-forward) outputs of layer~$\ell$.
Centering is linear, $h - \mu(h)\mathbf{1} = \sum_{m} \tilde{m}$ with $\tilde{m} = m - \mu(m)\mathbf{1}$, hence
\begin{equation}
  N(h) = \sum_{m \in \{e, a_1, f_1, a_2, f_2\}} \big\langle \gamma \odot \tilde{m},\, w_c \big\rangle .
\end{equation}
Each writer contributes additively to $N$, while $\sigma(h)$ depends on all writers jointly and scales every contribution alike.
The contribution of writer $m$ to the gain is
\begin{equation}
  s_m = \frac{\mathbb{E}^{(2)}\!\big[\langle \gamma \odot \tilde{m}, w_c \rangle / \sigma\big] - \mathbb{E}^{(1)}\!\big[\langle \gamma \odot \tilde{m}, w_c \rangle / \sigma\big]}{\mathbb{E}^{(2)}[A]} ,
  \qquad
  \sum_{m} s_m = 1 - \frac{1}{F} .
\end{equation}

\subsection{Gain factorization}
\label{app:gain-factorization}

For each context, 200 sequences of 20 triplets are generated, and the first triplet of each sequence is discarded.
The read-out is taken at every position whose next token is the parent and whose earlier tokens in the current triplet are children.
Positions with one child give $h^{(1)}$, positions with two value-concordant children give $h^{(2)}$, and positions with two value-discordant children enter only through the magnitude $\lvert v_c \rvert / \lambda_c$, with $\lambda_c = \log\big(p_c / (1 - p_c)\big)$.
All quantities are computed per model and context, and the models with the sparse layer in layer~1 and in layer~2 are pooled.

\subsection{Shrinkage in the variable plane}
\label{app:plane-shrinkage}

\paragraph{Drop.}
The normalization factor $F_\sigma$ exceeds one because the scale $\sigma$ of the residual is smaller after two value-concordant children than after one.
We measure this shrinkage by
\begin{equation}
  \Delta\sigma^2 = \mathbb{E}^{(1)}\big[\sigma^2\big] - \mathbb{E}^{(2)}\big[\sigma^2\big]
  = \tfrac{1}{d}\Big(\mathbb{E}^{(1)}\big[\lVert \tilde h \rVert^2\big] - \mathbb{E}^{(2)}\big[\lVert \tilde h \rVert^2\big]\Big) ,
\end{equation}
where $\tilde h = h - \mu(h)\mathbf{1}$.

\paragraph{Plane.}
We fit the regression of Appendix~\ref{app:functional-form} to the residual at the final LayerNorm input and take the variable plane as the span of the projections of the query term onto $d_{r_1}$ and $d_{r_2}$.
Let $\tilde h_{\parallel}$ be the orthogonal projection of $\tilde h$ onto this plane and $\tilde h_{\perp} = \tilde h - \tilde h_{\parallel}$. We write 
\begin{equation}
    \Delta\sigma^2 = \Delta\sigma^2_{\mathrm{plane}} + \Delta\sigma^2_{\mathrm{rest}},
\end{equation} where each term is
\begin{align}
  \Delta\sigma^2_{\mathrm{plane}} &= \tfrac{1}{d}\Big(\mathbb{E}^{(1)}\big[\lVert \tilde h_{\parallel} \rVert^2\big] - \mathbb{E}^{(2)}\big[\lVert \tilde h_{\parallel} \rVert^2\big]\Big)
  \\
  \Delta\sigma^2_{\mathrm{rest}} &= \tfrac{1}{d}\Big(\mathbb{E}^{(1)}\big[\lVert \tilde h_{\perp} \rVert^2\big] - \mathbb{E}^{(2)}\big[\lVert \tilde h_{\perp} \rVert^2\big]\Big).
\end{align}

The share of the plane is the sum of $\Delta\sigma^2_{\mathrm{plane}}$ over all models and contexts divided by the sum of $\Delta\sigma^2$.
The plane carries most of the drop. A cross-model baseline in comparison only captures a small fraction of the drop.

\subsection{Expansion of the normalization scale}
\label{app:sigma-expansion}

For writers $m, m'$ write $\mathrm{var}(m) = \tfrac{1}{d}\lVert \tilde m \rVert^2$ and $\mathrm{cov}(m, m') = \tfrac{1}{d}\langle \tilde m, \tilde m' \rangle$, the variance and covariance over the coordinates of the residual.
Since $\tilde h = \sum_{m} \tilde m$,
\begin{equation}
  \sigma^2(h) = \sum_{m} \mathrm{var}(m) + 2 \sum_{m < m'} \mathrm{cov}(m, m') + \epsilon .
\end{equation}

Applying $\mathbb{E}^{(1)} - \mathbb{E}^{(2)}$ term by term splits the drop exactly,
\begin{equation}
  \Delta\sigma^2 = \sum_{m} \Delta\mathrm{var}(m) + 2 \sum_{m < m'} \Delta\mathrm{cov}(m, m') ,
  \qquad
  \Delta(\cdot) = \mathbb{E}^{(1)}[\cdot] - \mathbb{E}^{(2)}[\cdot] .
\end{equation}

\paragraph{Held-out deficit.}
For each model, the held-out deficit of a term is its $\Delta$ in the held-out context minus the mean of its $\Delta$ in the two supervised contexts.
The fifteen term deficits sum to the deficit of $\Delta\sigma^2$.
The deficits of the variances are close to zero in both families, and the deficit is carried by the covariances (\ap{fig: term deficits}).
With the sparse layer in layer~1, it spreads over the covariances of $e$ and $a_1$ with the MLP writers.
The writers of the held-out context thus have the same magnitudes as in the supervised ones but are aligned differently, and no single pair accounts for the deficit.

\begin{figure}
    \centering
    \includegraphics[width=\linewidth]{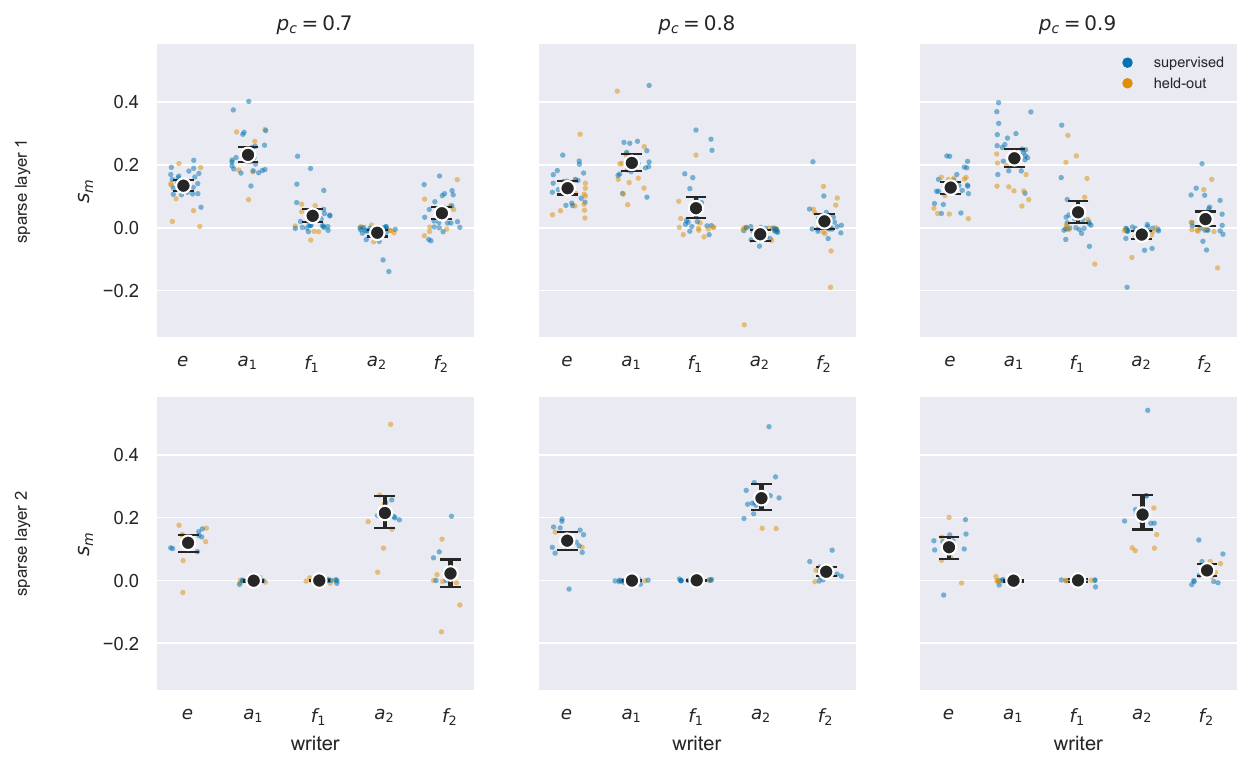}
    \caption{\textbf{Writers carrying the change in the readout.}
    For each writer $m \in \{e, a_1, f_1, a_2, f_2\}$, the share
    $s_m = \big(\mathbb{E}_2[\langle \gamma \odot \tilde m, w_c\rangle/\sigma]
          - \mathbb{E}_1[\langle \gamma \odot \tilde m, w_c\rangle/\sigma]\big)
          / \mathbb{E}_2[A]$
    is the writer's contribution to the change in the readout from one observed
    child ($\mathbb{E}_1$) to two concordant children ($\mathbb{E}_2$), normalized
    by the two-child readout $A = v - \langle b, w_c\rangle$. The layer-norm bias
    term $\langle b, w_c\rangle$ is constant and excluded. For each model and
    context, the five shares sum to $1 - 1/F$, where
    $F = \mathbb{E}_2[A]/\mathbb{E}_1[A]$ is the total readout gain.
    Columns: contexts $p_c$. Rows: models with sparse attention in layer~1
    (31 models) or layer~2 (14 models). Dots: individual models, blue for
    supervised contexts, orange for the held-out context. Black points: panel
    mean with 95\% percentile-bootstrap CIs (10{,}000 resamples). The attention
    writer of the sparse layer carries the largest share, followed by the
    embedding $e$. In layer-2 models, the layer-1 writers $a_1$ and $f_1$
    contribute almost nothing.}
    \label{fig:placeholder}
\end{figure}

\begin{figure}
    \centering
    \includegraphics[width=\linewidth]{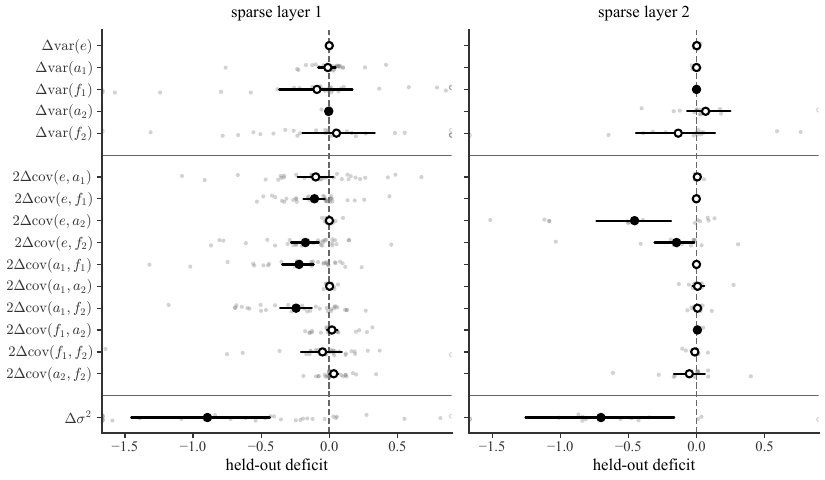}
    \caption{\textbf{Held-out shortfall in the drop of $\sigma^2$, by term.}
    For each model and term, $\Delta$ is the drop from one observed child to two
    concordant children ($\mathbb{E}_1 - \mathbb{E}_2$) in the held-out context,
    minus the mean drop over the model's two supervised contexts. Negative
    $\Delta$ means the held-out context shrinks the term less. Rows are the terms
    of the exact expansion of $\sigma^2$ over the five writers: five variances
    $\mathrm{var}(m) = \tfrac{1}{d}\lVert\tilde m\rVert^2$ and ten covariances
    $2\,\mathrm{cov}(m, m') = \tfrac{2}{d}\langle\tilde m, \tilde m'\rangle$. The
    bottom row is their sum, $\Delta\sigma^2$. Panels: models with sparse attention
    in layer~1 ($n = 31$) and layer~2 ($n = 14$). Grey dots are single models. Black
    points are means with 95\% percentile-bootstrap CIs over models (2{,}000
    resamples). A point is filled if its CI excludes zero and open otherwise. The x-axis spans the 2nd--98th
    percentile of all dots in both panels, padded by 5\%. The 24 of 720 dots beyond
    it are drawn open at the edge. In both families the held-out context shrinks
    $\sigma^2$ less. In layer-1 models the shortfall is spread over the covariances
    of $e$ and $a_1$ with the MLP writers $f_1$ and $f_2$. In layer-2 models it sits
    mainly in $2\Delta\mathrm{cov}(e, a_2)$ and $2\Delta\mathrm{cov}(e, f_2)$.}
    \label{fig:placeholder}
\end{figure}

\end{document}